\documentclass[a4paper,twoside]{article}

\usepackage{enumitem}
\usepackage{epsfig}
\usepackage{subcaption}
\usepackage{calc}
\usepackage{amssymb}
\usepackage{amstext}
\usepackage{amsmath}
\usepackage{amsthm}
\usepackage{multicol}
\usepackage{pslatex}
\usepackage{apalike}

\usepackage{graphicx}
\usepackage{amsmath,amssymb} 
\usepackage[normalem]{ulem}

\usepackage{multirow}
\usepackage[normalem]{ulem}
\useunder{\uline}{\ul}{}
\DeclareMathOperator{\softmax}{softmax}

\DeclareMathOperator{\alignf}{align}

\usepackage[activate]{microtype}
\newcommand{\hide}[1]{}

\newcommand{\MB}[1]{\mbox{\boldmath{$#1$}}} 

\newcommand{\open}[1]{\left(#1\right)} 
\newcommand{\W}[1]{\MB{W_{#1}}} 
\newcommand{\hi}{\MB{S}_{t}} 
\newcommand{\his}{\MB{\bar{S}}_{t}} 
\newcommand{\al}{\MB{a}_{n}} 

\newcommand{\tp}[1]{#1^\top} 

\usepackage{SCITEPRESS}     

\begin{document}

\title{An Enhanced Adversarial Network with Combined Latent Features for Spatio-Temporal Facial Affect Estimation in the Wild}

\author{\authorname{Decky Aspandi\sup{1,2}\orcidAuthor{0000-0002-6653-3470}, Federico Sukno\sup{1}\orcidAuthor{0000-0002-2029-1576}, Bj\"orn Schuller\sup{2,3}\orcidAuthor{0000-0002-6478-8699} and Xavier Binefa\sup{1}\orcidAuthor{0000-0002-4324-9952}}
\affiliation{\sup{1}Department of Information and Communication Technologies,  Pompeu Fabra University, Barcelona, Spain}
\affiliation{\sup{2}Chair of Embedded Intelligence for Health Care and Wellbeing, University of Augsburg, Germany}
\affiliation{\sup{3}GLAM -- Group on Language, Audio, \& Music, Imperial College London, UK}
\email{\{decky.aspandi, federico.sukno, xavier.binefa\}@upf.edu, schuller@informatik.uni-augsburg.de}
}


\abstract{Affective Computing has recently attracted the attention of the research community, due to its numerous applications in diverse areas. In this context, the emergence of video-based data allows to enrich the widely used spatial features with the inclusion of temporal information. However, such spatio-temporal modelling often results in very high-dimensional feature spaces and large volumes of data, making training difficult and time consuming. This paper addresses these shortcomings by proposing a novel model that efficiently extracts both spatial and temporal features of the data by means of its enhanced temporal modelling based on latent features. Our proposed model consists of three major networks, coined Generator, Discriminator, and Combiner, which are trained in an adversarial setting combined with curriculum learning to enable our adaptive attention modules. In our experiments, we show the effectiveness of our approach by reporting our competitive results on both the AFEW-VA and SEWA datasets, suggesting that temporal modelling improves the affect estimates both in qualitative and quantitative terms. Furthermore, we find that the inclusion of attention mechanisms leads to the highest accuracy improvements, as its weights seem to correlate well with the appearance of facial movements, both in terms of temporal localisation and intensity. Finally, we observe the sequence length of around 160\,ms to be the optimum one for temporal modelling, which is consistent with other relevant findings utilising similar lengths.}

\onecolumn \maketitle \normalsize \setcounter{footnote}{0} \vfill

\section{\uppercase{Introduction}}
\label{sec:introduction}

\noindent Affective Computing has recently attracted the attention of the research community, due to its numerous applications in diverse areas which include education~\cite{e_learning} or healthcare~\cite{Liu}, among others. The growing availability of affect-related datasets, such as AFEW-VA~\cite{afew} and the recently introduced SEWA~\cite{sewa} database enable the rapid development of deep learning-based techniques, which currently hold the state of the art. 

Further, the emergence of video-based data allows to enrich the widely used spatial features with the inclusion of temporal information. To this end, several authors have explored the use of long-short term memory (LSTM) recurrent neural networks (RNNs) \cite{temporallyAFEW,ma2019emotion}, endowed also with attention mechanisms~ \cite{luong2015effective,li2020exploring,xiaohua2019two}. However, such spatio-temporal modelling often results in very high-dimensional feature spaces and large volumes of data, making training difficult and time consuming. Moreover, it has been shown that the sequence length considered during training can be a decisive factor for successful temporal modelling \cite{afew,impactLength,farhadi2018re,CRCT}, and yet a detailed study of this aspect is lacking in the field.

This paper addresses both the lack of incorporation and analysis of temporal modelling on affective analysis. We propose a novel model which can be efficiently used to extract both spatial and temporal features of the data by means of its enhanced temporal modelling based on latent features. We do so by incorporating three major networks, coined Generator, Discriminator, and Combiner, which are trained in an adversarial setting to estimate the affect domains of Valence (V) and Arousal (A). Furthermore, we capitalise on these latent features to enable temporal modelling using LSTM RNNs, which we train progressively using curriculum learning enhanced with adaptive attention. Specifically, the contributions of this paper are as follows:

\begin{enumerate}[label=(\alph*)]
	\item We upgrade the standard adversarial setting, consisting of a Generator and a Discriminator, with a third network that combines the features from these networks, which are modified accordingly. This yields features that combine the latent space from the autoencoder-based Generator and a V-A Quadrant estimate produced by the modified Discriminator, resulting in a compact but meaningful representation that helps reduce the training complexity. 
	\item We propose the use of curriculum learning to enable analysis and optimisation of the temporal modelling length.
	\item We incorporate dynamic attention to further enhance our model estimates and show its effectiveness by reporting state of the art accuracy on both the AFEW-VA and SEWA datasets. 
\end{enumerate}

\section{\uppercase{Related Work}}

\noindent Affective Computing started by exploiting the use of classical machine learning techniques to enable  automatic affect estimation. Examples of early approaches include partial least squares regression~\cite{leastSqu}, and support vector machines~\cite{svr}. Subsequently, to further progress the investigations in this field, the development of larger and bigger datasets was addressed. Several datasets have been introduced so far, starting with SEMAINE~\cite{semaine}, AFEW-VA~\cite{afew}, RECOLA~\cite{recola}, OMG~\cite{barros2018omg}, AffectNet~\cite{mollahosseini2015affectnet}, and more recently SEWA~\cite{sewa}, aff-wild~\cite{affectChallenge,zafeiriou2017aff}, and~ aff-wild2\cite{kollias2019expression,kollias2020analysing}. Furthermore, the V-A labels have become the standard emotional dimensions over time, as opposed to hard emotion labels, given their continuous nature~\cite{afew,sewa}. \

Throughout the last few years, models based on Deep Learning have emerged and currently hold the state of the art in the context of affective analysis, given their ability to learn from large scale data. A recent example along this line is the work from Mitenkova et al.~\cite{mitenkova2019valence}, who introduce tensor modelling for affect estimations by using spatial features. In their work, they use tucker tensor regression optimised by means of deep gradient methods, thus allowing to preserve the structure of the data and reduce the number of parameters. Other works, such as \cite{simulAFEW}, adopt the multi-task approach to simultaneously address face detection and affective states prediction. Specifically, they use YOLO-based CNN models~\cite{huang2018yolo} to estimate the facial locations alongside V-A values through their proposed combined losses. As such, their models are able to incorporate the characteristics of facial attributes and estimate their relevance to affect inferences. \

\begin{figure*}[t]
  \centering
  \includegraphics[width=\textwidth]{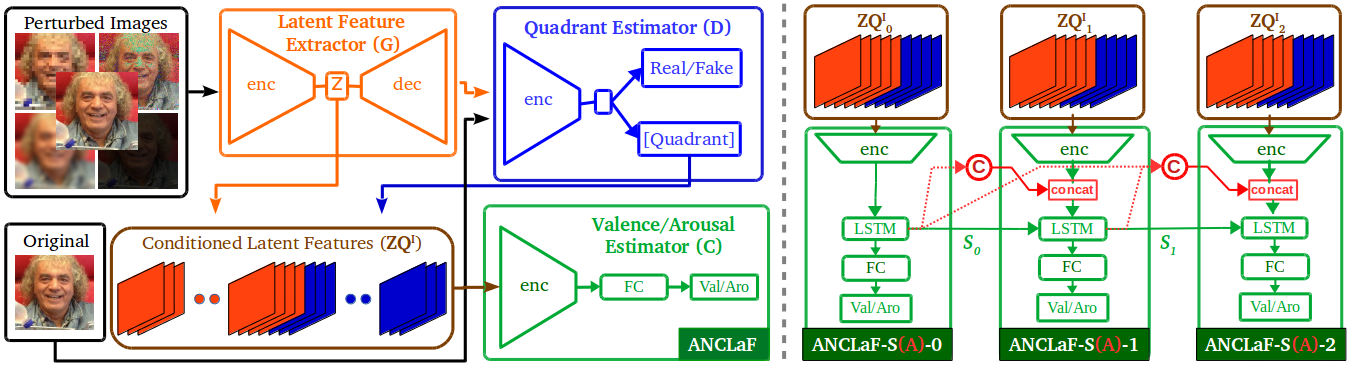}
  \caption{Schematic representation of our Full ANCLaF Networks. Left is our base model, which consists of three networks jointly trained in an adversarial setting: Latent Feature Extractor (G), Quadrant Estimator (D), and Valence Arousal Estimator (C). On the right, we see our network endowed with sequence modelling (ANCLaF-S) and attention mechanism (ANCLaF-SA). }
  \label{fig:overall}
\end{figure*}

The recent growth of video-based datasets has encouraged the inclusion of temporal modelling, which has shown to improve models'  training \cite{tempBenefit,tempImprove}. Relevant examples in Affective Computing include the works of Tellamekala et al.\ \cite{temporallyAFEW} and Ma et al.\ \cite{ma2019emotion}. In their work, Tellamekala et al.\ \cite{temporallyAFEW} enforce temporal coherency and smoothness aspects on their feature representation by constraining the differences between adjacent frames, while Ma et al.\ resort to the utilisation of LSTM RNNs with residual connections applied to multi-modal data. Furthermore, the use of attention has also been recently explored by Xiaohua et al.\ \cite{xiaohua2019two} and Li et al.\ \cite{li2020exploring}. Xiahoua et al.\ adopt multi-stage attention, which involves both spatial and temporal attention, on their facial based affect estimations. Meanwhile, using spectrogram data as input, Li et al.\ propose a deep network that utilises an attention mechanism \cite{luong2015effective} on top of their LSTM networks to predict the affective states.

Unfortunately, to our knowledge, all previous works involving temporal modelling on affective computing miss one important aspect of the analysis: the involved sequence length in their training. While the specified length of temporal modelling has been shown to affect the final results on other related facial analysis tasks \cite{afew,impactLength,farhadi2018re,CRCT}, the computational cost required to train large spatio-temporal models hampers one to address such analysis. However, these problems could be mitigated by: 1) the use of progressive sequence learning to permit step-wise observations of various sequence lengths; this approach has been shown in the recent work of \cite{CRCT} on facial landmark estimations, which uses curriculum learning enabling more robust training analysis and tuning of the temporal length; 2) the use of reduced feature sizes, enabling more efficient training process \cite{comas2019endtoend}; this has been explored in the affective computing field by the recent works such as \cite{aspandi2020adversarial}, which uses generative modelling to extract a latent space of representative features. These two aspects have inspired us to propose the combined models presented in this work, as explained in the next section.

\section{\uppercase{Methodology}}

\noindent Figure \ref{fig:overall} shows the overview of our proposed models, which consist of three networks: a Latent Feature Extractor (acting as Generator, G), a Quadrant Estimator (or Discriminator, D), and a Valence/Arousal Estimator (or Combiner, C). Given input image \textsc{I} which contains the facial area, both G and D will be responsible to learn low dimensional features that the combiner will use to estimate the associated Valence (V) and Arousal (A) state $\theta$. The architecture of both the G and D networks follows the recent work from \cite{aspandi2020adversarial}, and we propose to use LSTM enhanced with attention to create our C network. We proposed two main architecture variants: the \textbf{ANCLaF} network (left part of Figure \ref{fig:overall}), which uses single images as input and estimates V and A values 
independently for each frame, and \textbf{ANCLaF-S} and \textbf{ANCLaF-SA} (right part of Figure \ref{fig:overall}) that uses sequences of latent features extracted from $n$ frames as input, and utilises LSTM RNNs 
for the inference (\textbf{-S}), optionally combined with internal attention layers (\textbf{-SA}). 

\subsection{Adversarial Network with Combined Latent Features (ANCLaF)}
The pipeline of our base model \textbf{ANCLaF} starts with the G network. It receives either the original input image $\textsc{I}$, or a distorted version of it, $\tilde{I}$, as detailed in \cite{FADeNN,NoiseRedFamily}. It simultaneously produces the cleaned reconstruction of the input image $\hat{I}$ and a 2D latent representation that will be used as features ($\mathbb{Z}$): 
\begin{equation}
G(I)_{\varPhi^G} = dec_{\varPhi^G}(enc_{\varPhi^G}(I))\, with \, \mathbb{Z}^I \approx enc_{\varPhi^G}(I),
\end{equation}
\noindent where $\varPhi$ are the parameters of the respective networks, $enc$ and $dec$ are the encoder and decoder, respectively. Subsequently, the D network receives $\hat{I}$ and tries to estimate whether it was obtained from a true or fake example (namely, original or distorted input image), as well as a rough estimate of the affective state. In contrast with the formulation in \cite{aspandi2020adversarial}, in which D targets directly the intensity of V and A, we propose to base the estimated affect on the circumplex quadrant ($\mathbb{Q}$) \cite{russell1980circumplex} which discretises emotions along the valence and arousal dimensions (four quadrants). This, in turn, reduces the training complexity. Thus, letting FC stand for fully connected layer: 
\begin{equation}
D(I)_{\varPhi^D} = FC_{\varPhi^D}(enc_{\varPhi^D}(I)) \Rightarrow \mathbb{Q}^I \  and \  \{0,1\}.
\end{equation}
Then, $\mathbb{Q}$ is used to condition the extracted latent features $\mathbb{Z}$ through layer-wise concatenation, which we call $\mathbb{ZQ}$ \cite{cond_gan,ye2018channel}. Given these conditioned latent features, the C network performs the final stage of affect estimation, producing refined predictions of both V and A \cite{refine1,refine2,CRCT}. Thus, if $\hat{\theta}$ denote the estimated V and A:

\begin{equation}
    \begin{split}
    ANCLaF(I)&= C_{\varPhi^C}([G_{\varPhi^G}(I);D_{\varPhi^D}(G_{\varPhi^G}(I))]) \\
    &= C_{\varPhi^C}([\mathbb{Z}^I;\mathbb{Q}^I]) \\
    &= FC_{\varPhi^C}(enc_{\varPhi^C}([\mathbb{Z}^I;\mathbb{Q}^I])) \Rightarrow \hat{\theta}^I_{ANCLaF}.\\
    \end{split}
\end{equation}

\subsection{Attention Enhanced Sequence Latent Affect Networks}
We propose two sequence-based variants of our models: {ANCLaF-S} and {-SA}. Both of them use the combined features extracted by the G and D networks $\mathbb{ZQ}$, which are fed to LSTM networks to allow for temporal modelling \cite{LSTM} and complemented with an FC layer to produce the final estimates. These networks are trained using curriculum learning \cite{CL,farhadi2018re,CRCT}, in which the number of frames is progressively increased, allowing more 
throughout analysis of the training progress. Moreover, the training outcome for a given length facilitates the subsequent training of larger sequences \cite{farhadi2018re}. In this work, we considered a series of 2, 4, 8, 16, and 32 successive frames ($\textsc{N} = \{2,4,8,6,32\}$) for both training and inference stages. Depending on the number of frames to take into account (n), we use {ANCLaF-S-n} and {ANCLaF-SA-n} to name the respective variants of both {ANCLaF-S} and {ANCLaF-SA} networks.
Lastly, the main difference between the two sequence models is that {ANCLaF-SA} also includes internal attentional modelling using the current and previous internal states from the LSTM layers. Thus, V-A predictions of {ANCLaF-S-n} are:
\begin{multline}
    \forall n \in \textsc{N}\,,\, ANCLaF\textnormal{-}S\textnormal{-}n(I_n), h_n = \\ FC_{\varPhi^C}(LSTM_{\varPhi^C}([Z^I_n,Q^I_n],h_{n-1}))\\
    \Rightarrow FC_{\varPhi^C}(LSTM_{\varPhi^C}(\mathbb{ZQ}^I_n,h_{n-1})), 
\end{multline}

\noindent where LSTM is the Long Short Term Memory network \cite{LSTM}, and $h_n$ are  LSTM states (h) after n successive frames. Built upon $ANCLaF\textnormal{-}SA$, we further use attention modelling \cite{luong2015effective} to enable adaptive weights on model inferences by calculating the context vectors ($\mathbb{C}$) that summarise the importance of each previous state $h$. Differently from the original method, however, here, we also propose to include both the LSTM inner state (c) and outgoing states (h) \cite{concatenate} to provide the full previous information, and also to adapt these techniques to only consider n previous states following our curriculum learning approach. Hence, given the combined LSTM states at frame $t$, denoted ($S_t=[c_t,h_t]$), and $n$ previous states ($\bar{S}$), the alignment score is calculated as:
\begin{align}
\label{e:al}
\al(t)&=\alignf(\hi, \his)\,,\, with \, S_x = [h_x;c_x] \\
&=\frac{\exp \open{\MB{W_a} [\tp{\hi};\MB{\bar{S}}_{n}]}}{\sum_{N'} \exp \open{\MB{W_a} [\tp{\hi};\MB{\bar{S}}_{n'}]}}. \notag
\end{align}
Then, the location-based function computes the alignment scores from the previous states ($\bar{S}$): 
\begin{equation}
\al = \softmax(\W{a}\bar{S}).
\label{e:location}
\end{equation}
Given the alignment vector, it is used to compute the context vector $\mathbb{C}_t$ as the weighted average over the considered $n$ previous hidden states: 
\begin{equation}
\mathbb{C}_t = \frac{\sum_{n} a_n \odot S_n}{n}
\label{e:location}
\end{equation}
Finally, the context vector is concatenated with the current $\mathbb{ZQ}$ to be used as input to the C network pipeline: 
\begin{multline}
    \forall n \in \textsc{N}\,,\, ANCLaF\textnormal{-}SA\textnormal{-}n(I_n), h_n = \\ FC_{\varPhi^C}(LSTM_{\varPhi^C}([\mathbb{C}_n;\mathbb{ZQ}^I_n],h_{n-1})).
\end{multline}

\subsection{Training Losses}
We use the modified adversarial training from \cite{aspandi2020adversarial} to train both the G and D networks, and incorporate the training of the C network by providing the latter with the features extracted from both the G and D nets on the fly. With this setup, we allow C to benefit from the improved quality of the features extracted by G and D as their training progresses. The equations for the modified adversarial training of these three networks are: 
\begin{equation}
\begin{split}
\mathcal{L}_{adv} = & \thinspace {\mathbb{E}}_{I} \left[ \log{{D}(I)} \right]  \> \>  +   \\
& \thinspace {\mathbb{E}}_{I}[\log{(1 - {D}(G(\tilde{I})))} + 
  \thinspace {\mathbb{E}}_{afc}[C(I), \theta_I].
\end{split}
\label{eq1}
\end{equation}
We use
similar ${L}_{afc}$ losses as in \cite{aspandi2020adversarial}, which incorporates multiple affect metrics: Rooted Mean Square Error (RMSE) (Eq.\ \ref{mse}), Correlation(COR) (Eq.\ \ref{cor}), Concordance Correlation Coefficients (CCC) (Eq.\ \ref{icc}), and \cite{afew} with the addition of Intra-class Correlation Coefficient (ICC)\cite{sewa}. Thus, with \{$\hat{\theta}$,$\theta$\} as the predicted and the ground truth V-A values, the $\mathcal{L}_{afc}$ is defined as follows:
\begin{equation}
\label{afc}
\mathbb{E}_{afc} = \sum_{i = 1}^{F} \frac{f_i}{F} (\mathcal{L}_{RMSE} + \mathcal{L}_{COR} + \mathcal{L}_{CCC} + \mathcal{L}_{ICC})
\end{equation}
\begin{equation}
\label{mse}
\mathcal{L}_{RMSE} = \sqrt{\frac{1}{n} \sum_{i=1}^{n} (\hat{\theta_i}-\theta_i)^2}, \\ 
\end{equation}
\begin{equation}
\label{cor}
\mathcal{L}_{COR} = \frac{\mathbb{E}[(\hat{\theta}-\hat{\mu_{\theta}}) - (\theta-\mu_{\theta})]}{\sigma_{\hat{\theta}}\sigma_{\theta}}  \\ \end{equation}
\begin{equation}
\label{icc}
\mathcal{L}_{CCC} =  2x\frac{\mathbb{E}[(\hat{\theta}-\hat{\mu_{\theta}}) - (\theta-\mu_{\theta})]}{\sigma_{\hat{\theta}}^2+\sigma_{\theta}^2  + (\mu_{\hat{\theta}}-\mu_{\theta})^2}
\end{equation}
\begin{equation}
\label{=icc}
\mathcal{L}_{ICC} =  2x\frac{\mathbb{E}[(\hat{\theta}-\hat{\mu_{\theta}}) - (\theta-\mu_{\theta})]}
{\sigma_{\hat{\theta}}^2+\sigma_{\theta}^2 },
\end{equation}

\noindent where $f_i$ is the total number of instances of discrete V-A classes $i$, and $F$ is a normalisation factor \cite{NoiseRedFamily} for the total V-A classes (discretised by a value of 10). This normalisation factor is crucial in cases of large imbalance in the number of instances per class, like in the AFEW-VA dataset (see Section \ref{sec:expSet}). \

\subsection{Model Training} 
We use both the AFEW~\cite{afew} and SEWA~\cite{sewa} datasets to train all our model variants, by following their original subject-independent protocol (5-fold cross validation). 
We conducted two training stages for each of our proposed models. Firstly, we trained the G, D, and C networks simultaneously using adversarial loss as indicated in Equation \ref{eq1}. This training stage produced our baseline results without any sequential modelling, and conditional latent features $\mathbb{ZQ}$ to be used for the next stages of ANCLaF-S(A) Training. \

In the second stage, The training of both ANCLaF-S and ANCLaF-SA was performed using the combined latent and quadrant features, under the previously defined curriculum learning scheme. We progressively train our ANCLaF-S models from 2, 4, 8, 16 to 32 steps of temporal modelling with multi-stage transfer learning \cite{christodoulidis2016multisource}. Subsequently, we add our proposed attention mechanism to the pre-trained ANCLaF-S models, thus obtaining our ANCLaF-SA models. In both cases, we optimise the affect loss defined in Equation \ref{afc} with the same experimental settings used to train ANCLaF. 

We need to note that our combined training setup translates to more than 100 experiments in total. Hence, the use of latent features (known as a good choice to achieve reduced dimensionality representations) is critical to speed up our training process and make our experiments feasible. We observed a saving up to 1 : 4 of the original times during training each of our models by using the extracted latent features, with respect to using the original image (around 12 hours versus 2 days) on a single NVIDIA Titan X GPU. Full definitions of our models can be found in the respective online source code\footnote{https://github.com/deckyal/Seq-Att-Affect}.  

\section{\uppercase{Experiments and Results}}
\subsection{Datasets and Experiment Settings}
\label{sec:expSet}
To quantify the impact of our temporal modelling, we opted to use two of the most popular and accessible video datasets available: Acted Facial Expressions in the Wild (AFEW-VA)\cite{afew} and Automatic Sentiment Analysis in the Wild (SEWA)\cite{sewa}. On the one hand, AFEW-VA has more individual examples (600 versus 538) than SEWA, however, the latter has more frame examples, more contextual information (such as subject, id of the associated culture) and is more balanced in terms of V-A labels \cite{mitenkova2019valence}. 
Furthermore, both datasets contain \emph{in the wild} situations, enabling real time model evaluations. Finally, the labels provided are in the form of continuous V-A values, together with additional facial landmark locations that we refined further using other external models \cite{CRCT} to obtain more stable detection of the facial area.

In each experiment, we provide the results from all variants of our models to highlight the contribution of each module: first, we evaluate the ANCLaF model, which operates by exclusively using the latent features extracted on each frame ($\mathbb{ZQ}$) without any temporal modelling.  Then, we provide results from both ANCLaF-S and ANCLaF-SA, which incorporate temporal modelling (and attention in the case of -SA). We report both RMSE and COR results, on both datasets, adding also ICC and CCC metrics for the AFEW-VA and SEWA datasets, respectively, to facilitate quantitative analysis to other results reported in the literature. Finally, for fair comparisons, we compare our models against external results which followed similar experimental protocols, i.\,e., using exclusively this dataset in their training stage.

\begin{figure*}[t!]
  \centering
  \includegraphics[width=\linewidth]{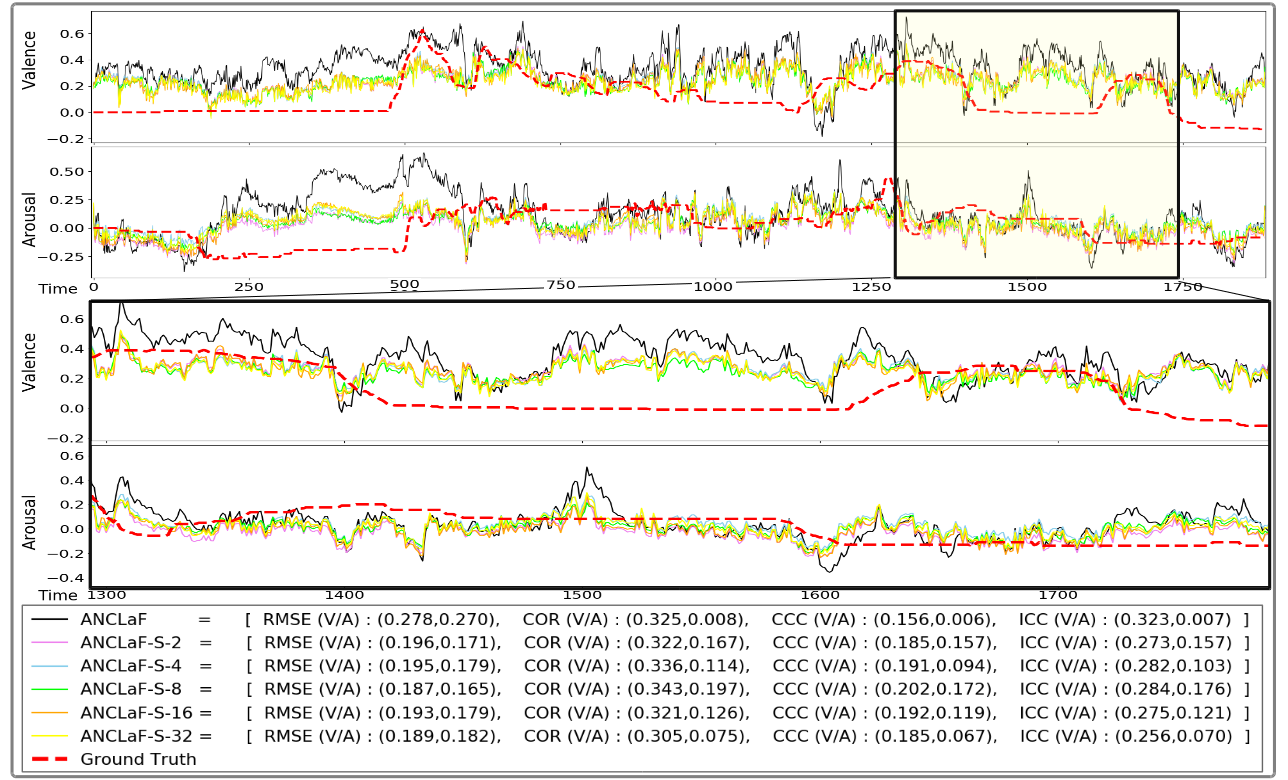}
  \caption{Analysis of prediction results from a single frame (ANCLaF) and from multiple frames with temporal modelling (ANCLaF-S-n). Top: the overview of the overall results; Bottom:, a closer look at the prediction results.}
  \label{fig:nseq}
\end{figure*}

\subsection{Comparative Results}
\label{sec:comparativeRes}
Table \ref{tab:afew} and table \ref{tab:sewa} provide the full comparisons of our proposed models against other reported results for both the AFEW-VA and SEWA datasets, respectively. We can identify several findings based on these results: Firstly, that our base ANCLaF model, relying on a single image at a time, can produce quite competitive accuracy compared to other results from the literature. Furthermore, its accuracy is also higher than the results from the original AEG-CD-SZ models in which it is based upon \cite{aspandi2020adversarial}, as shown by its higher accuracy on the SEWA datasets, especially for Valence. This may indicate its better processing capabilities of the visual features, considering that AEG-CD-SZ also incorporates audio features, which in a way also explains its higher accuracy on the prediction of Arousal.

\setlength{\tabcolsep}{1.5pt}
\begin{table}[h]
\begin{center}
\caption{Quantitative comparisons on the AFEW-VA dataset.}
\def\arraystretch{1}
\resizebox{\columnwidth}{!}{%
\begin{tabular}{lccccccccc}
\hline  \hline
\multicolumn{1}{c}{\multirow{2}{*}{Model}} & \multicolumn{3}{c}{RMSE$\;\downarrow$ } & \multicolumn{3}{c}{COR$\;\uparrow$ } & \multicolumn{3}{c}{ICC$\;\uparrow$ } \\ \cline{2-10} 
\multicolumn{1}{c}{} & \multicolumn{1}{c}{VAL} & \multicolumn{1}{c}{ARO} & \multicolumn{1}{c}{AVG} & 
\multicolumn{1}{c}{VAL} & \multicolumn{1}{c}{ARO} & \multicolumn{1}{c}{AVG} & 
\multicolumn{1}{c}{VAL} & \multicolumn{1}{c}{ARO} & \multicolumn{1}{c}{AVG} \\ \hline\hline
Baseline \cite{afew} & 2.680 & 2.275 & 2.478 & \textbf{0.407} & 0.450 & \textbf{0.429} & 0.290 & 0.356 & 0.323 \\ \hline
Coherent\cite{temporallyAFEW} & - & - & - & 0.293 & 0.426 & 0.360 & - & - &  \\ \hline
Simul \cite{simulAFEW}& 2.600 & 2.500 & 2.550 & 0.390 & 0.290 & 0.340 & \textbf{0.320} & 0.210 & 0.265 \\ \hline \hline
ANCLaF & 2.682 & 2.344 & 2.513 & 0.306 & 0.399 & 0.353 & 0.219 & 0.309 & 0.264 \\ \hline \hline
ANCLaF-S-2 & 2.675 & 2.295 & 2.485 & 0.314 & 0.410 & 0.362 & 0.236 & 0.296 & 0.266 \\ \hline
ANCLaF-S-4 & 2.654 & 2.279 & 2.467 & 0.303 & 0.420 & 0.361 & 0.224 & 0.307 & 0.266 \\ \hline
ANCLaF-S-8 & 2.595 & 2.202 & 2.398 & 0.328 & 0.425 & 0.377 & 0.272 & 0.344 & 0.308 \\ \hline
ANCLaF-S-16 & 2.617 & 2.292 & 2.454 & 0.302 & 0.401 & 0.351 & 0.224 & 0.299 & 0.261 \\ \hline
ANCLaF-S-32 & 2.568 & 2.328 & 2.448 & 0.288 & 0.405 & 0.346 & 0.214 & 0.304 & 0.259 \\ \hline
ANCLaF-S-AVG & 2.622 & 2.279 & 2.450 & 0.307 & 0.412 & 0.360 & 0.234 & 0.310 & 0.272 \\ \hline \hline
ANCLaF-SA-2 & 2.540 & 2.241 & 2.390 & 0.373 & 0.454 & 0.413 & 0.291 & 0.353 & 0.322 \\ \hline
ANCLaF-SA-4 & 2.586 & 2.260 & 2.423 & 0.386 & 0.445 & 0.415 & 0.302 & 0.342 & 0.322 \\ \hline
ANCLaF-SA-8 & \textbf{2.481} & 2.239 & \textbf{2.360} & 0.371 & \textbf{0.467} & 0.419 & 0.294 & \textbf{0.367} & \textbf{0.331} \\ \hline
ANCLaF-SA-16 & 2.601 & \textbf{2.225} & 2.413 & 0.377 & 0.467 & 0.422 & 0.294 & 0.363 & 0.328 \\ \hline
ANCLaF-SA-32 & 2.581 & 2.256 & 2.419 & 0.361 & 0.436 & 0.399 & 0.270 & 0.332 & 0.301 \\ \hline
ANCLaF-SA-AVG & 2.558 & 2.244 & 2.401 & 0.373 & 0.454 & 0.414 & 0.290 & 0.352 & 0.321 \\ \hline \hline
\end{tabular}%
}
\label{tab:afew}
\end{center}
\end{table}

\setlength{\tabcolsep}{1.5pt}
\begin{table}[h]
\begin{center}
\caption{Quantitative comparisons on the SEWA dataset.}
\def\arraystretch{1}
\resizebox{\columnwidth}{!}{%
\begin{tabular}{llccccccccc}
\hline \hline
\multicolumn{1}{c}{\multirow{2}{*}{Model}} & \multicolumn{3}{c}{RMSE $\;\downarrow $}& \multicolumn{3}{c}{COR $\;\uparrow$ } & \multicolumn{3}{c}{CCC $\;\uparrow$ } \\ \cline{2-10} 
\multicolumn{1}{c}{} & \multicolumn{1}{c}{VAL} & \multicolumn{1}{c}{ARO} & \multicolumn{1}{c}{AVG} & 
\multicolumn{1}{c}{VAL} & \multicolumn{1}{c}{ARO} & \multicolumn{1}{c}{AVG} & 
\multicolumn{1}{c}{VAL} & \multicolumn{1}{c}{ARO} & \multicolumn{1}{c}{AVG} \\ \hline\hline
Baseline \cite{sewa} & - & - & - & 0.350 & 0.350 & 0.350 & 0.350 & 0.290 & 0.320 \\ \hline
Tensor \cite{mitenkova2019valence}  & 0.334 & 0.380 & 0.357 & 0.503 & 0.439 & 0.471 & 0.469 & 0.392 & 0.431 \\ \hline
AEG-CD-SZ\cite{aspandi2020adversarial}  & \textbf{0.323} & 0.350 & 0.337 & 0.442 & \textbf{0.478} & 0.460 & 0.405 & \textbf{0.430} & 0.418 \\ \hline \hline
ANCLaF & 0.354 & 0.347 & 0.351 & 0.530 & 0.395 & 0.462 & 0.492 & 0.364 & 0.428 \\ \hline \hline
ANCLaF-S-2 & 0.349 & 0.345 & 0.347 & 0.533 & 0.396 & 0.464 & 0.503 & 0.368 & 0.436 \\ \hline
ANCLaF-S-4 & 0.344 & 0.336 & 0.340 & 0.536 & 0.403 & 0.469 & 0.510 & 0.382 & 0.446 \\ \hline
ANCLaF-S-8 & 0.341 & 0.339 & 0.340 & 0.538 & 0.404 & 0.471 & 0.514 & 0.381 & 0.448 \\ \hline
ANCLaF-S-16 & 0.354 & 0.344 & 0.349 & 0.527 & 0.395 & 0.461 & 0.490 & 0.369 & 0.429 \\ \hline
ANCLaF-S-32 & 0.353 & 0.346 & 0.349 & 0.527 & 0.396 & 0.461 & 0.494 & 0.368 & 0.431 \\ \hline
ANCLaF-S-AVG & 0.348 & 0.342 & 0.345 & 0.532 & 0.399 & 0.465 & 0.502 & 0.374 & 0.438 \\ \hline \hline
ANCLaF-SA-2 & 0.343 & 0.333 & 0.338 & 0.545 & 0.420 & 0.482 & 0.509 & 0.390 & 0.449 \\ \hline
ANCLaF-SA-4 & 0.336 & \textbf{0.328} & \textbf{0.332} & 0.550 & 0.429 & 0.490 & 0.526 & 0.399 & 0.463 \\ \hline
ANCLaF-SA-8 & 0.336 & 0.332 & 0.334 & \textbf{0.558} & 0.424 & \textbf{0.491} & \textbf{0.529} & 0.405 & \textbf{0.467} \\ \hline
ANCLaF-SA-16 & 0.334 & 0.331 & \textbf{0.332} & 0.556 & 0.421 & 0.488 & 0.528 & 0.393 & 0.461 \\ \hline
ANCLaF-SA-32 & 0.336 & 0.362 & 0.349 & 0.550 & 0.418 & 0.484 & 0.513 & 0.389 & 0.451 \\ \hline
ANCLaF-SA-AVG & 0.337 & 0.337 & 0.337 & 0.552 & 0.422 & 0.488 & 0.521 & 0.395 & 0.458 \\ \hline \hline
\end{tabular}%
}
\label{tab:sewa}
\end{center}
\end{table}
\begin{figure*}[t!]
  \centering
  \includegraphics[width=.9\linewidth]{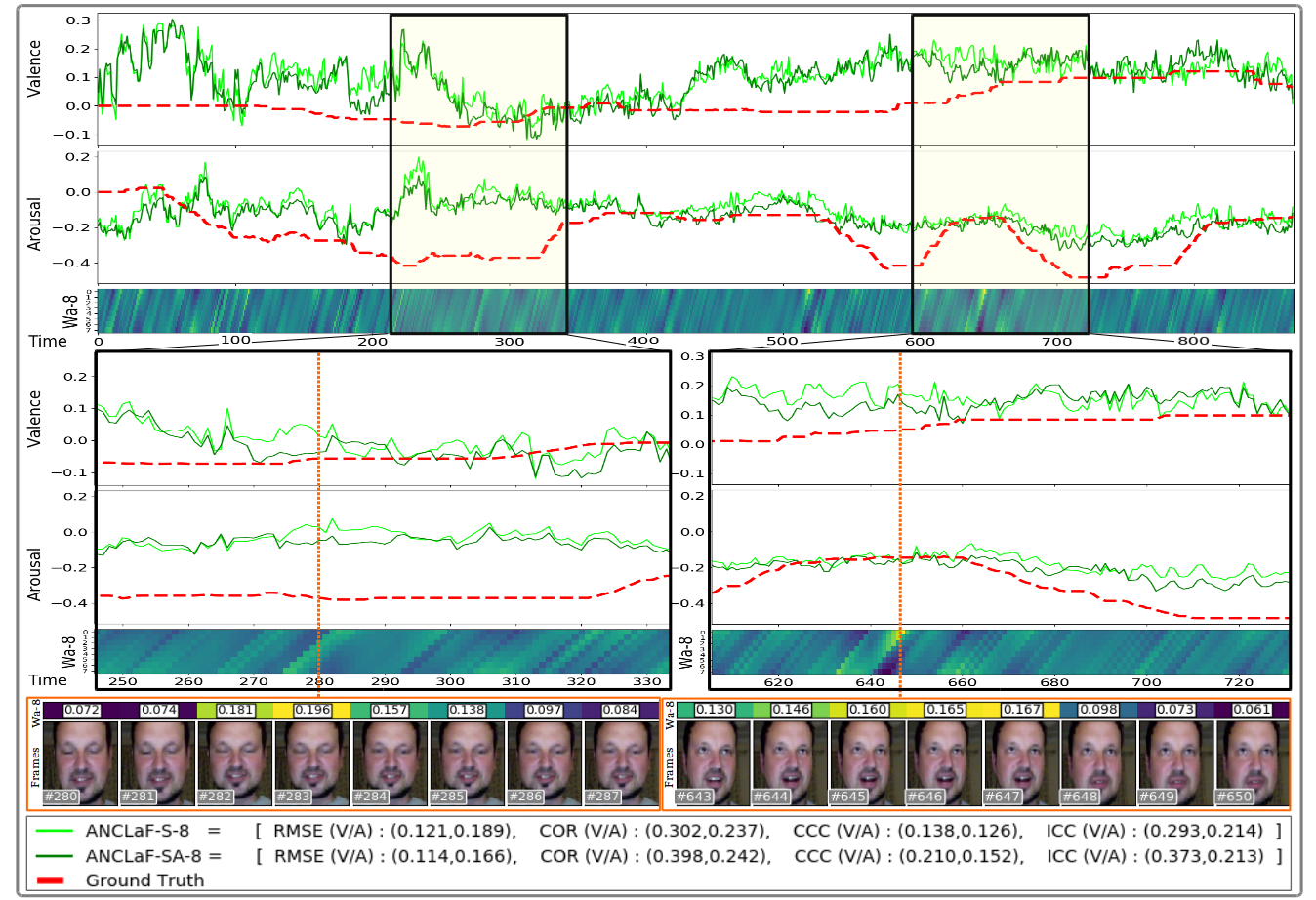}
  \caption{Analysis of the attention impact on the prediction results of our sequence modelling (results from ANCLaF-S-8 and ANCLaF-SA-8, which correspond to the best ANCLaF-S and ANCLaF-SA models, respectively). Top: overview of the overall results; Bottom: two examples of a closer view on the prediction graph. The column Wa-8 shows the attention weights learnt for the eight considered frames.}
  \label{fig:nat}
\end{figure*}

Secondly, we notice a slight accuracy improvement when our models incorporate sequence modelling  (ANCLaF-S), especially in terms of correlations, namely, concordeance corelation coefficient (CCC), and 
ICC. This finding demonstrates the benefit of the temporal modelling, yielding more stable results than those achieved by ANCLaF (cf.\  Section \ref{sec:analysisSeq}). However, even though the overall accuracy of ANCLaF-S is better than that of ANCLaF (and quite comparable to other state of the art models), the improvement can be considered modest, especially if we compare it with the improvement achieved when we include attention in our models. 
Indeed, we can see that our ANCLaF-SA outperforms almost all compared models across the different affect metrics. These findings suggest that the plain utilisation of LSTMs may not be enough to attain a considerable and substantial increase of accuracy  \cite{schmitt2019continuous}, justifying the inclusion of the attention mechanism in our approach.


Thirdly, we further observe a noticeable trend of steady increase in accuracy from the predictions of both ANCLaF-S and ANCLaF-SA as the number of considered frames grows from 2 to 8, and then it plateaus (or even worsens a bit) as $n$ continues to increase. This trend suggests that generally, a medium sequence length (between 4 to 16 frames) is optimal to produce more accurate predictions and that too short and too long sequences degrade temporal modelling. This finding is quite consistent with those from \cite{CRCT}, indicating the importance of progressive learning, which allows us to analyse and choose the optimal sequence length during training. Lastly, this sequence length selection may also impact the context vector along with its weights learnt in our attentional module, which explains why a similar trend was observed in the results from these models (see Section \ref{sec:analysisAtt} for more details).

\subsection{Analysis of the Impact of Sequence Modelling}
\label{sec:analysisSeq}
Figure \ref{fig:nseq} shows an example of V-A predictions for ANCLaF and ANCLaF-S-n, together with the ground-truth annotations. Specifically, in the top part, we can see the predicted affect states from our models that, in general, are quite related to the ground truth values. However, we notice that the results of our sequence based models are more accurate than their non-sequential counterparts. We can also see that the the predicted values from ANCLaF are quite sparse, thus,  quite unstable compared to ANCLAF-S, which explains its lower 
COR, CCC, and ICC values. Our sequence modelling, on the other hand, is able to create smooth predictions with higher overall accuracy. 

On the bottom part of the figure, showing a magnified portion of the same example, we further notice that the results for all ANCLaF-S-n are quite similar, with those from ANCLaF-S-8 showing the highest resemblance to the ground-truth. Thus, inclusion of too short or too long sequences yields sub-optimal results due to the complexity of the facial movements included between frames (see the next section for further details).


\begin{figure*}[t!]
  \centering
  \includegraphics[width=\linewidth]{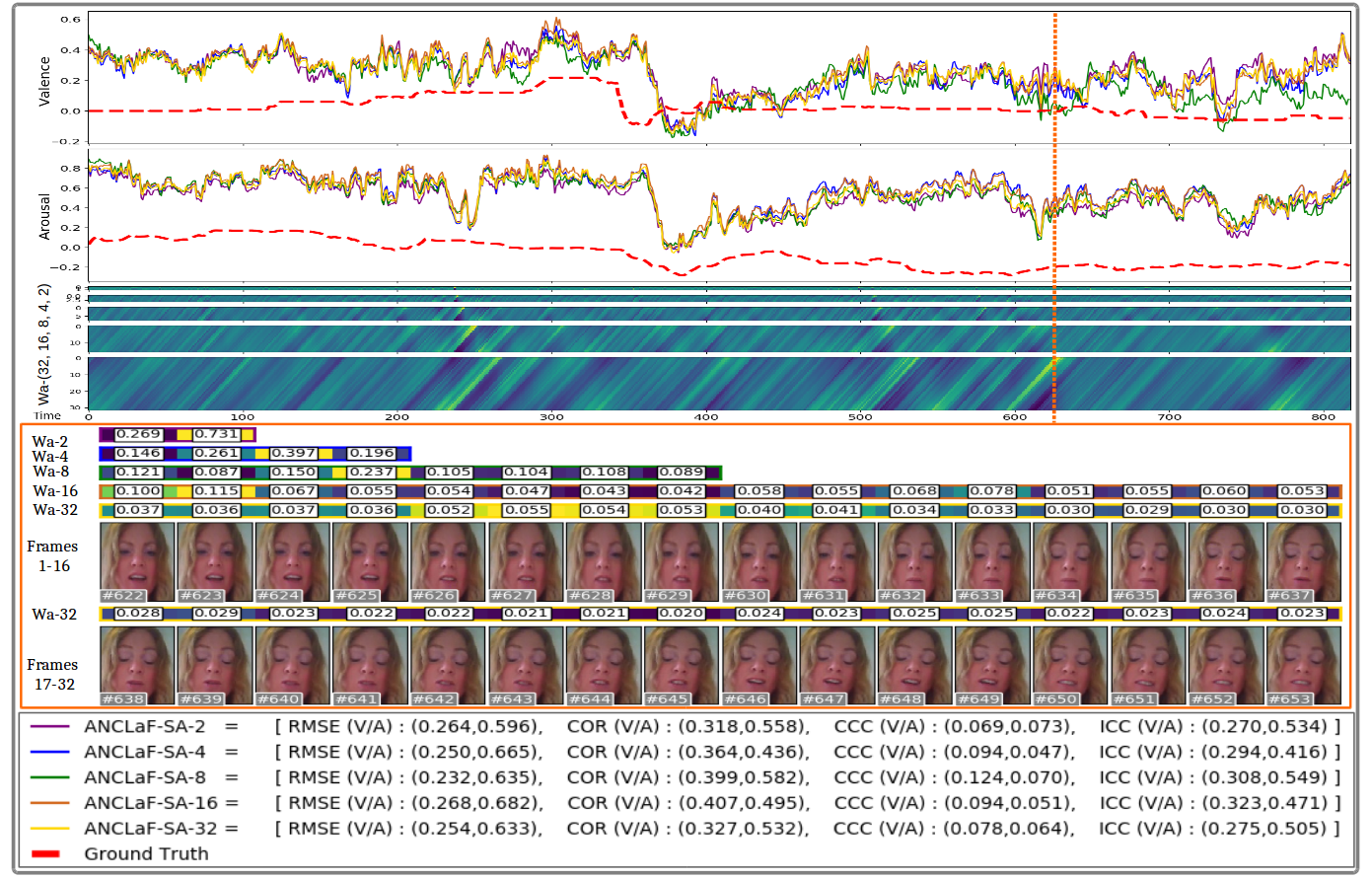}
  \caption{Analysis of the relationship between the selection of sequence length (n) and the learnt weights of our attentional approach. Top: overview of the prediction results of all variants of our models with attention mechanism (ANCLaF-SA-n) alongside their learnt weights. Middle: details for frames 622 to 653 with their associated weights for each model. Bottom: legend containing the quantitative comparisons.}
  \label{fig:seqat}
\end{figure*}

\subsection{Analysis of the Role of the Learnt Attentions Weights}
\label{sec:analysisAtt}

To analyse the impact of the attention mechanism on our sequence modelling, we first show in Figure \ref{fig:nat} a comparison of our baseline sequence modelling (ANCLaF-S) against ANCLaF-SA with attention activated. In the top part, we can see the predictions from the best performing models with and without attention (ANCLaF-S-8 and ANCLaF-SA-8). Comparing the predictions from both models, we find that the results are quite similar, though in some cases ANCLaF-SA seems to be more accurate and closer to the ground truth. The quantitative accuracy results indicated on the respective legends confirm this observation. 

The attention weights learnt by ANCLaF-SA, involving the previous eight frames, are also displayed at the bottom of the prediction plots. We can see that the weights calculated with respect to the associated frames seem to be higher in the presence of changes. Indeed, we observe that the attention weights are usually activated prior to subsequent facial movements. Interestingly, the intensity of the activations also appears to highlight the level of these facial movements, or the changes between frames. For instance, from frames 280-287, we can see that the different level of the weight intensity seems to be small, which also correlates to the subtle changes observed in those frames (e.\,g., closing of the eyes). In contrast, in frames 643-650, we see high levels of activation on the first few frames that correspond to the more discernible facial movements on the respective frames, such as the changes observed in the mouth area. These correlations illustrate how our models are capable of learning temporal changes.


Figure \ref{fig:seqat} provides further details on the attention mechanism for different temporal modelling lengths. We can see that all the displayed models show quite smooth results, thanks to the temporal modelling, but not all of them achieve the same accuracy on the predictions. The bottom part of the figure, highlighting the input sequence from frames 622 to 653, can help to provide an intuition about the optimal temporal modelling length, which was found to be about $8$ frames. To this end, let us start by looking at the whole set of 32 frames: we can see that such a  sequence of frames comprises multiple facial changes, and considering all of them together makes the training task harder to optimise. On the other hand, if we consider groups of very few frames (e.\,g., 2 or 4 frames), the system is likely to capture only part of a given facial action, which may impede it to properly interpret it. Therefore, we see that the optimal sequence length is the one that contains enough frames to interpret facial changes without extending too much the temporal context, which may unnecessarily increase training complexity and reduce accuracy. 

Finally, it is important to emphasise that the optimal sequence length needs to take into account the frame rate and the specific facial movements that are present in each dataset. In the considered dataset, with an overall frame rates of 50\,fps, this length corresponds to 160\,ms. 


\section{\uppercase{Conclusions}}
\noindent In this work, we have successfully built a sequence-attention based neural network for affect estimations in the wild. We did so by incorporating three major sub-networks: the Generator, which is responsible to extract latent features on each frame; the Discriminator, which is used to supply the first step of affect estimates of emotional quadrant, and the Combiner, which merges latent features and quadrant information to produce the final refined affect estimates of Valence and Arousal on a frame by frame basis. We then added an LSTM layer to allow temporal modelling, which we further enhanced by using step-wise attention modelling. We trained these three major sub-networks in an adversarial setting, and used curriculum learning on the sequential training stages.

We showed the effectiveness of our approach by reporting top state of the art results on two of the most widely used video datasets for affect analysis, namely AFEW-VA and SEWA. Specifically, our baseline models, which operate without any sequence modelling, yield quite competitive results with other models reported in the literature. On the other hand, our more advanced models, which are sequence-based, clearly helped to improve the affect estimates both in qualitative and quantitative terms. Qualitatively, the temporal modelling helped to produce more stable results, with visibly smoother transitions between affect predictions. Quantitatively, our models produced the overall best accuracy results reported so far on both tested datasets. 

Within sequence-based models, we observed the highest accuracy improvements when the attention mechanism was included. Detailed analysis of the attention weights highlighted their correlation with the appearance of facial movements, both in terms of (temporal) localisation and intensity. Finally, we found a sequence length of around 160\,ms to be the optimum one for temporal modelling, which is consistent with other relevant findings utilising similar lengths.

Future work will need to explore further optimisation of the considered adversarial topologies and attention mechanisms as well as their transferability across databases, cultures, and domains. 

\section*{\uppercase{Acknowledgments}}
\noindent This work is partly supported by the Spanish Ministry of Economy and Competitiveness under project grant TIN2017-90124-P, the Maria de Maeztu Units of Excellence Programme (MDM-2015-0502), and the donation bahi2018-19 to the CMTech at UPF. Further funding has been received from the European Union's Horizon 2020 research and innovation programme under grant agreement No.\ 826506 (sustAGE).

\bibliographystyle{apalike}
{\small
\bibliography{example}}

\end{document}